\documentclass[11pt,a4paper]{article}
\usepackage[hyperref]{acl2019}
\usepackage{times}
\usepackage{latexsym}

\usepackage{url}

\usepackage{tabularx}
\usepackage{multirow}
\usepackage{graphicx}
\usepackage{subfigure}
\usepackage{amsmath}
\usepackage{amssymb}
\usepackage{booktabs}
\usepackage{color, colortbl}
\definecolor{LightCyan}{rgb}{0.88,1,1}
\definecolor{Gray}{gray}{0.9}

\usepackage{tikz}

\aclfinalcopy 
\def\aclpaperid{2526} 

\title{Matching Article Pairs with Graphical Decomposition and Convolutions}

\author{Bang Liu$^\dagger$, Di Niu$^\dagger$, Haojie Wei$^\ddagger$, Jinghong Lin$^\ddagger$, Yancheng He$^\ddagger$, Kunfeng Lai$^\ddagger$, Yu Xu$^\ddagger$
   \\ $^\dagger$University of Alberta, Edmonton, AB, Canada
   \\ \texttt{\{bang3, dniu\}@ualberta.ca}
   \\ $^\ddagger$Platform and Content Group, Tencent, Shenzhen, China
   \\ \texttt{\{fayewei, daphnelin, collinhe, calvinlai, henrysxu\}@tencent.com}}

\date{}

\begin{document}
\maketitle
\begin{abstract}
Identifying the relationship between two articles, e.g., whether two articles published from different sources describe the same breaking news, is critical to many document understanding tasks.
Existing approaches for modeling and matching sentence pairs do not perform well in matching \emph{longer} documents, which embody more complex interactions between the enclosed entities than a sentence does.
To model article pairs, we propose the \textit{Concept Interaction Graph} to represent an article as a graph of concepts. We then match a pair of articles by comparing the sentences that enclose the same concept vertex through a series of encoding techniques, and aggregate the matching signals through a graph convolutional network. 
To facilitate the evaluation of long article matching, we have created two datasets, each consisting of about 30K pairs of \emph{breaking news} articles covering diverse topics in the open domain.
Extensive evaluations of the proposed methods on the two datasets demonstrate significant improvements over a wide range of state-of-the-art methods for natural language matching.
\end{abstract}

\section{Introduction}
\label{sec:intro}

Identifying the relationship between a pair of articles is an essential natural language understanding task, which is critical to news systems and search engines. For example, a news system needs to cluster various articles on the Internet reporting the same breaking news (probably in different ways of wording and narratives), remove redundancy and form storylines \cite{shahaf2013information,liu2017growing,zhou2015unsupervised,vossen2015storylines,bruggermann2016storyline}.  The rich semantic and logic structures in longer documents have made it a different and more challenging task to match a pair of articles than to match a pair of sentences or a query-document pair in information retrieval.


Traditional term-based matching approaches estimate the semantic distance between a pair of text objects via unsupervised metrics, e.g., via TF-IDF vectors, BM25 \cite{robertson2009probabilistic}, LDA \cite{blei2003latent} and so forth. These methods have achieved success in query-document matching, information retrieval and search.
In recent years, a wide variety of deep neural network models have also been proposed for text matching \cite{hu2014convolutional,qiu2015convolutional,wan2016deep,pang2016text}, which can capture the semantic dependencies (especially sequential dependencies) in natural language through layers of recurrent or convolutional neural networks.
However, existing deep models are mainly designed for matching sentence pairs, e.g., for paraphrase identification, answer selection in question-answering, omitting the complex interactions among keywords, entities or sentences that are present in a longer article. Therefore, article pair matching remains under-explored in spite of its importance. 

In this paper, we apply the divide-and-conquer philosophy to matching a pair of articles and bring deep text understanding from the currently dominating sequential modeling of language elements to a new level of graphical document representation, which is more suitable for longer articles. Specifically, we have made the following contributions:

\textit{First}, 
we propose the so-called \textit{Concept Interaction Graph} (CIG) to represent a document as a weighted graph of concepts, where each concept vertex is either a keyword or a set of tightly connected keywords. The sentences in the article associated with each concept serve as the features for local comparison to the same concept appearing in another article. Furthermore, two concept vertices in an article are also connected by a weighted edge which indicates their interaction strength. 
The CIG does not only capture the essential semantic units in a document but also offers a way to perform anchored comparison between two articles along the common concepts found.


\textit{Second}, we propose a divide-and-conquer framework to match a pair of articles based on the constructed CIGs and graph convolutional networks (GCNs). The idea is that for each concept vertex that appears in both articles, we first obtain the local matching vectors through a range of text pair encoding schemes, including both neural encoding and term-based encoding. We then aggregate the local matching vectors into the final matching result through graph convolutional layers \cite{kipf2016semi, defferrard2016convolutional}. In contrast to RNN-based sequential modeling, our model factorizes the matching process into local matching sub-problems on a graph, each focusing on a different concept, and by using GCN layers, generates matching results based on a holistic view of the entire graph.




Although there exist many datasets for sentence matching, the semantic matching between longer articles is a largely unexplored area. To the best of our knowledge, to date, there does not exist a labeled public dataset for long document matching. To facilitate evaluation and further research on document and especially news article matching, we have created \emph{two labeled datasets}\footnote{Our code and datasets are available at: https://github.com/BangLiu/ArticlePairMatching}, one annotating whether two news articles found on Internet (from different media sources) report the same breaking news event, while the other annotating whether they belong to the same news story (yet not necessarily reporting the same breaking news event). These articles were collected from major Internet news providers in China, including Tencent, Sina, WeChat, Sohu, etc., covering diverse topics, and were labeled by professional editors.
Note that similar to most other natural language matching models, all the approaches proposed in this paper can easily work on other languages as well.

Through extensive experiments, we show that our proposed algorithms have achieved significant improvements on matching news article pairs, as compared to a wide range of state-of-the-art methods, including both term-based and deep text matching algorithms. 
With the same encoding or term-based feature representation of a pair of articles, our approach based on graphical decomposition and convolutions can improve the classification accuracy by $17.31\%$ and $23.09\%$ on the two datasets, respectively.

\section{Concept Interaction Graph}
\label{sec:cig}

\begin{figure}[tb]
\centering
\includegraphics[width=3.0in]{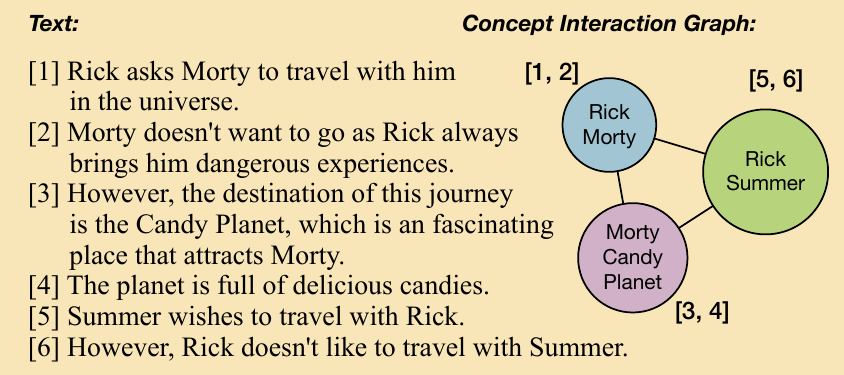}
\vspace{-3mm}
\caption{An example to show a piece of text and its  Concept Interaction Graph representation.}
\label{fig:CaseStudy}
\vspace{-5mm}
\end{figure}

\begin{figure*}
\includegraphics[width=\textwidth]{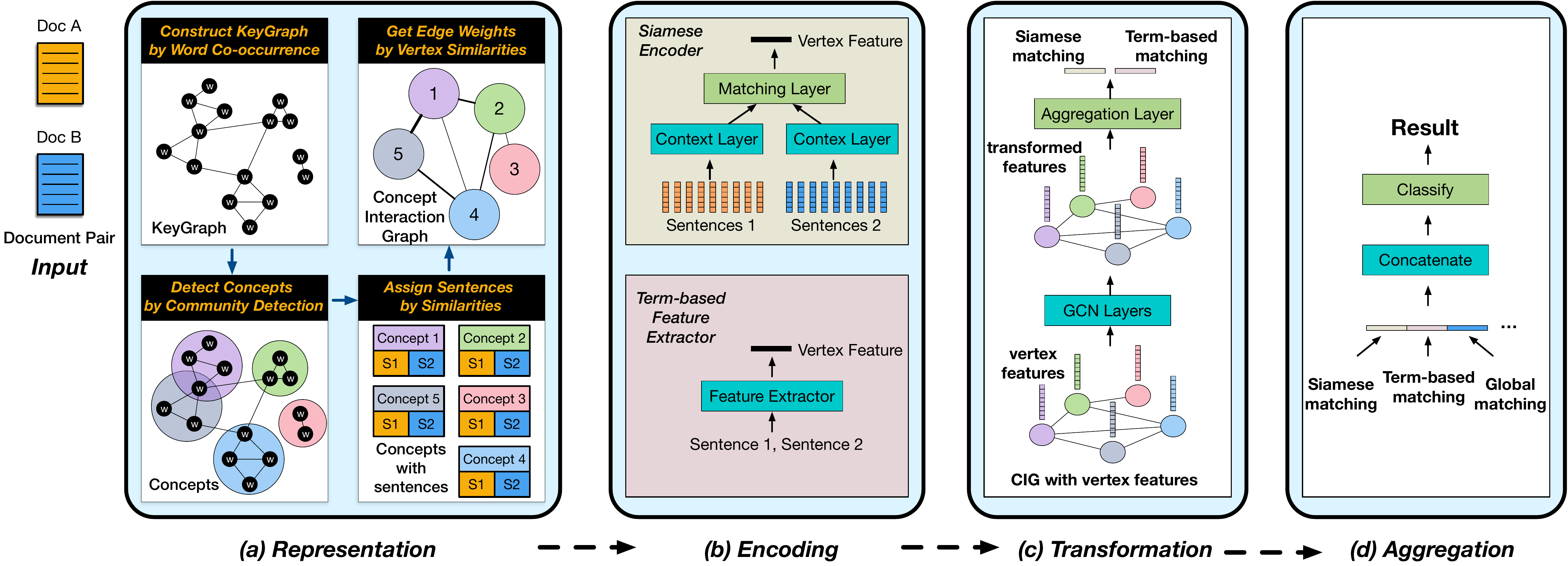}
\vspace{-3mm}
\caption{An overview of our approach for constructing the \textit{Concept Interaction Graph} (CIG) from a pair of documents and classifying it by Graph Convolutional Networks.}
\label{fig:System}
\vspace{-4mm}
\end{figure*}

In this section, we present our \textit{Concept Interaction Graph} (CIG) to represent a document as an \emph{undirected weighted graph}, which decomposes a document into subsets of sentences, each subset focusing on a different \emph{concept}. 
Given a document $\mathcal{D}$, a CIG is a graph $G_D$,
where each vertex in $G_D$ is called a \textit{concept}, which is a keyword or a set of highly correlated keywords in document $\mathcal{D}$.
Each sentence in $\mathcal{D}$ will be attached to the \emph{single} concept vertex that it is the most related to, which most frequently is the concept the sentence mentions. Hence, vertices will have their own sentence sets, which are disjoint.
The weight of the edge between a pair of concepts denotes how much the two concepts are related to each other and can be determined in various ways.

As an example, Fig.~\ref{fig:CaseStudy} illustrates how we convert a document into a Concept Interaction Graph. We can extract keywords \textit{Rick}, \textit{Morty}, \textit{Summer}, and \textit{Candy Planet} from the document using standard keyword extraction algorithms, e.g., TextRank \cite{mihalcea2004textrank}. These keywords are further clustered into three concepts, where each concept is a subset of highly correlated keywords. After grouping keywords into concepts, we attach each sentence in the document to its most related concept vertex. For example, in Fig.~\ref{fig:CaseStudy}, sentences $1$ and $2$ are mainly talking about the relationship between \textit{Rick} and \textit{Morty}, and are thus attached to the concept (\textit{Rick}, \textit{Morty}). Other sentences are attached to vertices in a similar way. The attachment of sentences to concepts naturally dissects the original document into multiple disjoint sentence subsets. As a result, we have represented the original document with a graph of key concepts, each with a sentence subset, as well as the interaction topology among them.


Fig~\ref{fig:System} (a) illustrates the construction of CIGs for a pair of documents aligned by the discovered concepts. Here we first describe the detailed steps to construct a CIG for a single document:

\textbf{KeyGraph Construction.}
Given a document $\mathcal{D}$, we first extract the named entities and keywords by TextRank~\cite{mihalcea2004textrank}.
After that, we construct a keyword co-occurrence graph, called \emph{KeyGraph}, based on the set of found keywords. Each keyword is a vertex in the KeyGraph. We connect two keywords by an edge if they co-occur in a same sentence.

We can further improve our model by performing co-reference resolution and synonym analysis to merge keywords with the same meaning. However, we do not apply these operations due to the time complexity.

\textbf{Concept Detection (Optional).}
The structure of KeyGraph reveals the connections between keywords. If a subset of keywords are highly correlated, they will form a densely connected sub-graph in the KeyGraph, which we call a \textit{concept}.
Concepts can be extracted by applying community detection algorithms on the constructed KeyGraph. Community detection is able to split a KeyGraph ${G}_{\text{key}}$ into a set of communities $C = \{\mathcal{C}_1, \mathcal{C}_2, ..., \mathcal{C}_{|C|}\}$, where each community $\mathcal{C}_i$  contains the keywords for a certain concept. 
By using overlapping community detection, each keyword may appear in multiple concepts. 
As the number of concepts in different documents varies a lot, we utilize the \emph{betweenness centrality score} based algorithm \cite{sayyadi2013graph} to detect keyword communities in KeyGraph.

Note that this step is optional, i.e., we can also use each keyword directly as a concept. The benefit brought by concept detection is that it reduces the number of vertices in a graph and speeds up matching, as will be shown in Sec.~\ref{sec:eval}.

\textbf{Sentence Attachment.}
After the concepts are discovered, the next step is to group sentences by concepts.
We calculate the cosine similarity between each sentence and each concept, where sentences and concepts are represented by TF-IDF vectors. We assign each sentence to the concept which is the most similar to the sentence.
Sentences that do not match any concepts in the document will be attached to a \emph{dummy vertex} that does not contain any keywords.

\textbf{Edge Construction}.
To construct edges that reveal the correlations between different concepts, for each vertex, we represent its sentence set as a concatenation of the sentences attached to it, and calculate the edge weight between any two vertices as
the TF-IDF similarity between their sentence sets.
Although edge weights may be decided in other ways, our experience shows that constructing edges by TF-IDF similarity generates 
a CIG that is more densely connected.

When performing article pair matching, the above steps will be applied to a pair of documents $\mathcal{D}_A$ and $\mathcal{D}_B$, as is shown in Fig.~\ref{fig:System} (a). The only additional step is that we align the CIGs of the two articles by the concept vertices, and for each common concept vertex, merge the sentence sets from $\mathcal{D}_A$ and $\mathcal{D}_B$ for local comparison.

\section{Article Pair Matching through Graph Convolutions}
\label{sec:textmatch}

Given the merged CIG $G_{AB}$ of two documents $\mathcal{D}_A$ and $\mathcal{D}_B$ described in Sec.~\ref{sec:cig}, we match a pair of articles in a ``divide-and-conquer'' manner by matching the sentence sets from $\mathcal{D}_A$ and $\mathcal{D}_B$ associated with each concept and aggregating local matching results into a final result through multiple graph convolutional layers.  
Our approach overcomes the limitation of previous text matching algorithms, by extending text representation from a sequential (or grid) point of view to a graphical view, and can therefore better capture the rich semantic interactions in longer text.

Fig.~\ref{fig:System} illustrates the overall architecture of our proposed method, which consists of four steps: a) representing a pair of documents by a single merged CIG, b) learning multi-viewed matching features for each concept vertex, c) structurally transforming local matching features by graph convolutional layers, and d) aggregating local matching features to get the final result. Steps (b)-(d) can be trained end-to-end.



\textbf{Encoding Local Matching Vectors}.
Given the merged CIG $G_{AB}$, our first step is to learn an appropriate \emph{matching vector} of a fixed length for each individual concept $v\in G_{AB}$ to express the semantic similarity between 
$\mathcal S_A(v)$ and $\mathcal S_B(v)$, the sentence sets of concept $v$ from documents $\mathcal{D}_A$ and $\mathcal{D}_B$, respectively. This way, the matching of two documents is converted to match the pair of sentence sets on each vertex of $G_{AB}$.
Specifically, we generate local matching vectors based on both neural networks and term-based techniques.

\textit{Siamese Encoder}: we apply a Siamese neural network encoder~\cite{neculoiu2016learning} onto each vertex $v\in G_{AB}$ to convert the word embeddings \cite{mikolov2013efficient} of $\{\mathcal S_A(v), \mathcal S_B(v)\}$ into a fixed-sized hidden feature vector $\mathbf m_{AB}(v)$, which we call the \emph{match vector}.

We use a Siamese structure to take $\mathcal S_A(v)$ and $\mathcal S_B(v)\}$ (which are two sequences of word embeddings) as inputs, and
encode them into two context vectors through the context layers that share the same weights, as shown in Fig.~\ref{fig:System} (b). The context layer usually contains one or multiple bi-directional LSTM (BiLSTM) or CNN layers with max pooling layers, aiming to capture the contextual information in $\mathcal S_A(v)$ and $\mathcal S_B(v)\}$.

Let $\mathbf{c}_A(v)$ and $\mathbf{c}_B(v)$ denote the context vectors obtained for $\mathcal S_A(v)$ and $\mathcal S_B(v)$, respectively. Then, the matching vector $\mathbf{m}_{AB}(v)$ for vertex $v$ is given by the subsequent aggregation layer, which concatenates the element-wise absolute difference and the element-wise multiplication of the two context vectors, i.e., 
\begin{equation}
\mathbf{m}_{AB}(v) = (|\mathbf{c}_A(v) - \mathbf{c}_B(v)|, \mathbf{c}_A(v) \circ \mathbf{c}_B(v)) ,
\end{equation}
where $\circ$ denotes Hadamard product.

\textit{Term-based Similarities}:
we also generate another matching vector for each $v$ by directly calculating term-based similarities between $\mathcal S_A(v)$ and $\mathcal S_B(v)$, based on 5 metrics: the TF-IDF cosine similarity, TF cosine similarity, BM25 cosine similarity, Jaccard similarity of 1-gram, and Ochiai similarity measure. These similarity scores  are concatenated into another matching vector $\mathbf{m}'_{AB}(v)$ for $v$, as shown in Fig.~\ref{fig:System} (b).


\textbf{Matching Aggregation via GCN}
The local matching vectors must be aggregated into a final matching score for the pair of articles. We propose to utilize the ability of the Graph Convolutional Network (GCN) filters~\cite{kipf2016semi} to capture the patterns exhibited in the CIG $G_{AB}$ at multiple scales. 
In general, the input to the GCN is a graph $G = (\mathcal{V}, E)$ with $N$ vertices $v_i \in \mathcal{V}$, and edges $e_{ij} = (v_i, v_j) \in E$ with weights $w_{ij}$. The input also contains a vertex feature matrix denoted by $X = \{\mathbf x_i\}_{i=1}^{N}$, where $\mathbf x_i$ is the \emph{feature vector} of vertex $v_i$. 
For a pair of documents $\mathcal{D}_A$ and $\mathcal{D}_B$, we input their CIG $G_{AB}$ (with $N$ vertices) with a (concatenated) matching vector on each vertex into the GCN, such that the feature vector of vertex $v_i$ in GCN
is given by \[\mathbf x_i = (\mathbf m_{AB}(v_i), \mathbf m'_{AB}(v_i)).\] 

Now let us briefly describe the GCN layers~\cite{kipf2016semi} used in Fig.~\ref{fig:System} (c).
Denote the weighted adjacency matrix of the graph as $A \in \mathbb{R}^{N\times N}$ where $A_{ij} = w_{ij}$ (in CIG, it is the TF-IDF similarity between vertex $i$ and $j$). Let $D$ be a diagonal matrix such that $D_{ii} = \sum_j A_{ij}$. The input layer to the GCN is $H^{(0)} = X$, which contains the original vertex features. Let $H^{(l)} \in \mathbb{R}^{N\times M_l}$ denote the matrix of \emph{hidden representations} of the vertices in the $l^{\text{th}}$ layer.
Then each GCN layer applies the following graph convolutional filter onto the previous hidden representations:
\begin{equation}
\label{eq:gcn}
H^{(l+1)} = \sigma(\tilde{D}^{-\frac{1}{2}} \tilde{A} \tilde{D}^{-\frac{1}{2}} H^{(l)} W^{(l)}) ,
\end{equation}
where $\tilde{A} = A + I_N$,  $I_N$ is the identity matrix, and $\tilde{D}$ is a diagonal matrix such that $\tilde{D}_{ii} = \sum_j \tilde{A_{ij}}$. They are the adjacency matrix and the degree matrix of graph $G$, respectively.

$W^{(l)}$ is the trainable weight matrix in the $l^{\text{th}}$ layer. $\sigma(\cdot)$ denotes an activation function such as sigmoid or ReLU function. Such a graph convolutional rule is motivated by the first-order approximation of localized spectral filters on graphs~\cite{kipf2016semi} and when applied recursively, can extract interaction patterns among vertices.

Finally, the hidden representations in the final GCN layer is merged into a single vector (called a graphically merged matching vector) of a fixed length, denoted by $\mathbf m_{AB}$, by 
taking the mean of the hidden vectors of all vertices in the last layer. The final matching score will be computed based on $\mathbf m_{AB}$, through a classification network, e.g., a multi-layered perceptron (MLP).

In addition to the graphically merged matching vector $\mathbf m_{AB}$ described above, we may also append other global matching features to $\mathbf m_{AB}$ to expand the feature set. These additional global features can be calculated, e.g., by encoding two documents directly with state-of-the-art language models like BERT \cite{devlin2018bert} or by directly computing their term-based similarities. However, we show in Sec.~\ref{sec:eval} that such global features can hardly bring any more benefit to our scheme, as the graphically merged matching vectors are already sufficiently expressive in our problem.


\section{Evaluation}
\label{sec:eval}

\begin{figure}
\centering
\includegraphics[width=3.0in]{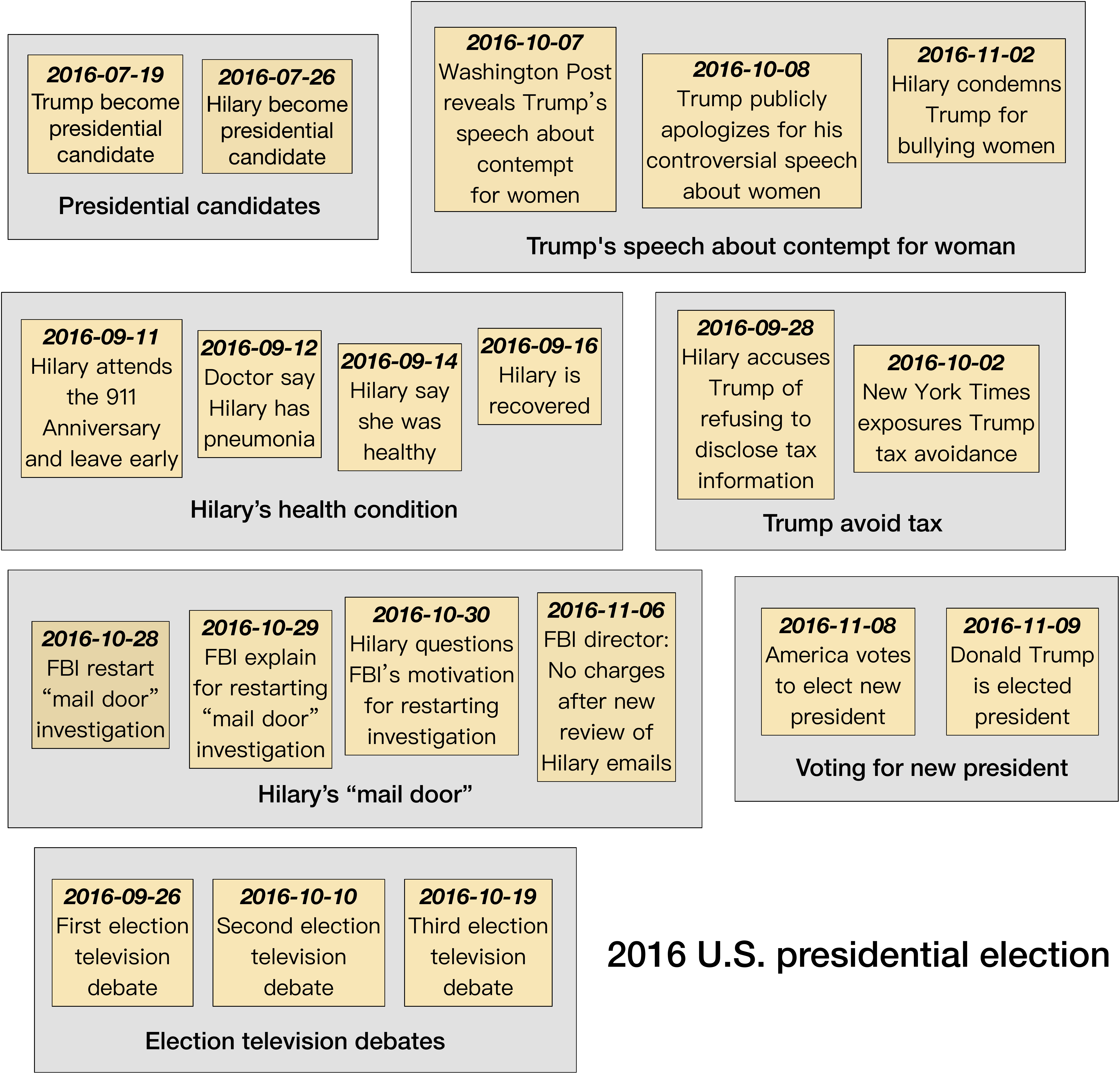}
\vspace{-7mm}
\caption{The events contained in the story ``2016 U.S. presidential election''.}
\label{fig:caseStudy-event-story}
\vspace{-6mm}
\end{figure}



\textbf{Tasks}.
We evaluate the proposed approach on the task of \textbf{\emph {identifying whether a pair of news articles report the same breaking news (or event) and whether they belong to the same series of news story}}, which is motivated by a real-world news app. In fact, the proposed article pair matching schemes have been deployed in the anonymous news app for news clustering, with more than $110$ millions of daily active users.

Note that traditional methods to document clustering include unsupervised text clustering and text classification into predefined topics. However, a number of breaking news articles emerge on the Internet everyday with their topics/themes unknown, so it is not possible to predefine their topics. Thus, supervised text classification cannot be used here. It is even impossible to determine how many news clusters there exist. Therefore, the task of classifying whether two news articles are reporting the same breaking news event or belong to the same story is critical to news apps and search engines for clustering, redundancy removal and topic summarization.

In our task, an ``event'' refers to a piece of breaking news on which multiple media sources may publish articles with different narratives and wording.  Furthermore, a ``story'' consists of a series of logically related breaking news events. 
It is worth noting that our objective is fundamentally different from the traditional event coreference literature, e.g., \cite{bejan2010unsupervised,lee2013deterministic,lee2012joint} or SemEval-2018 Task 5 (Counting Events) \cite{postma2018semeval}, where the task is to 
detect all the events (or in fact, ``actions'' like shooting, car crashes) a document mentions.

In contrast, although a news article may mention multiple entities and even previous physical events, the ``event'' in our dataset always refers to the breaking news that the article intends to report or the incident that triggers the media's coverage.
And our task is to identify whether two articles intend to report the same breaking news. For example, two articles ``University of California system libraries break off negotiations with Elsevier, will no longer order their journals'' and ``University of California Boycotts Publishing Giant Elsevier'' from two different sources are apparently intended to report the same breaking news event of UC dropping subscription to Elsevier, although other actions may be peripherally mentioned in these articles, e.g., ``eight months of unsuccessful negotiations.'' In addition,
we do not attempt to perform reading comprehension question answering tasks either, e.g., finding out how many killing incidents or car crashes there are in a year (SemEval-2018 Task 5 \cite{postma2018semeval}).

As a typical example, Fig.~\ref{fig:caseStudy-event-story} shows the events contained in the story \textit{2016 U.S. presidential election}, where each tag shows a breaking news event possibly reported by multiple articles with different narratives (articles not shown here). We group highly coherent events together. For example, there are multiple events about Election television debates. One of our objectives is to identify whether two news articles report the same event, e.g., a yes when they are both reporting \textit{Trump and Hilary's first television debate}, though with different wording, or a no, when one article is reporting \textit{Trump and Hilary's second television debate} while the other is talking about \textit{Donald Trump is elected president}.

\textbf{Datasets}.
To the best of our knowledge, there is no publicly available dataset for long document matching tasks. We created two datasets: the Chinese News Same Event dataset (CNSE) and Chinese News Same Story dataset (CNSS), which are labeled by professional editors. They contain long Chinese news articles collected from major Internet news providers in China, covering diverse topics in the open domain.
The CNSE dataset contains $29,063$ pairs of news articles with labels representing whether a pair of news articles are reporting about the same breaking news event.
Similarly, the CNSS dataset contains $33,503$ pairs of articles with labels representing whether two documents fall into the same news story.
The average number of words for all documents in the datasets is $734$ and the maximum value is $21791$.

In our datasets, we only labeled the \emph{major} event (or story) that a news article is reporting, since in the real world, each breaking news article on the Internet must be intended to report some specific breaking news that has just happened to attract clicks and views. Our objective is to determine whether two news articles intend to report the same breaking news. 

Note that the negative samples in the two datasets are not randomly generated: we select document pairs that contain similar keywords, and exclude samples with TF-IDF similarity below a certain threshold. The datasets have been made publicly available for research purpose.

\begin{table}[tb]
\tabcolsep=0.11cm
\small
  \caption{Description of evaluation datasets.}
  \label{tab:datasets}
  \begin{tabular}{llllll}
    \toprule
    Dataset & Pos Samples & Neg Samples & Train & Dev & Test\\
    \midrule
    CNSE & $12865$ & $16198$ & $17438$ & $5813$ & $5812$ \\
    CNSS & $16887$ & $16616$ & $20102$ & $6701$ & $6700$ \\
    \bottomrule
  \end{tabular}
  \vspace{-7mm}
\end{table}

Table~\ref{tab:datasets} shows a detailed breakdown of the two datasets. 
For both datasets, we use $60\%$ of all the samples as the training set, $20\%$ as the development (validation) set, and the remaining $20\%$ as the test set. We carefully ensure that different splits do not contain any overlaps to avoid data leakage.
The metrics used for performance evaluation are the accuracy and F1 scores of binary classification results. For each evaluated method, we perform training for 10 epochs and then choose the epoch with the best validation performance to be evaluated on the test set.

\begin{table*}[tb]
\centering
\tabcolsep=0.11cm
\small
  \caption{Accuracy and F1-score results of different algorithms on CNSE and CNSS datasets.}
  \label{tab:experiments}
  \resizebox{\textwidth}{!}{
  \begin{tabular}{l|cc|cc!{\vrule width 1pt}l|cc|cc}
    \toprule
    \multirow{2}{*}{Baselines} & \multicolumn{2}{c}{CNSE} & \multicolumn{2}{c!{\vrule width 1pt}}{CNSS} & \multirow{2}{*}{Our models} & \multicolumn{2}{c}{CNSE} & \multicolumn{2}{c}{CNSS} \\ 
    & Acc & F1 & Acc & F1 & & Acc & F1 & Acc & F1\\
    \midrule
     \textbf{I}. ARC-I & $53.84$ & $48.68$ & $50.10$ & $66.58$ & \textbf{XI}. CIG-Siam & $74.47$ & $73.03$ & $75.32$ & $78.58$ \\
    \textbf{II}. ARC-II & $54.37$ & $36.77$ & $52.00$ & $53.83$ & \textbf{XII}. CIG-Siam-GCN & $\mathbf{74.58}$ & $\mathbf{73.69}$ & $\mathbf{78.91}$ & $\mathbf{80.72}$ \\
    \textbf{III}. DUET & $55.63$ & $51.94$ & $52.33$ & $60.67$  & \textbf{XIII}. CIG$_{cd}$-Siam-GCN  & $73.25$ & $73.10$ & $76.23$ & $76.94$ \\  \cline{6-10}
    \textbf{IV}. DSSM & $58.08$ & $64.68$ & $61.09$ & $70.58$  & \textbf{XIV}. CIG-Sim & $72.58$ & $71.91$ & $75.16$ & $77.27$ \\
    \textbf{V}. C-DSSM & $60.17$ & $48.57$ & $52.96$ & $56.75$  & \textbf{XV}. CIG-Sim-GCN & $\mathbf{83.35}$ & $\mathbf{80.96}$ & $\mathbf{87.12}$ & $\mathbf{87.57}$ \\
    \textbf{VI}. MatchPyramid & $66.36$ & $54.01$ & $62.52$ & $64.56$  & \textbf{XVI}. CIG$_{cd}$-Sim-GCN & $81.33$ & $78.88$ & $86.67$ & $87.00$ \\\cline{1-10}
    \textbf{VII}. BM25 & $69.63$ & $66.60$ & $67.77$ & $70.40$  & \textbf{XVII}. CIG-Sim\&Siam-GCN & $84.64$ & $\mathbf{82.75}$ & $89.77$ & $90.07$ \\
    \textbf{VIII}. LDA & $63.81$ & $62.44$ & $62.98$ & $69.11$  & \textbf{XVIII}. CIG-Sim\&Siam-GCN-Sim$^{g}$ & $84.21$ & $82.46$ & $\mathbf{90.03}$ & $\mathbf{90.29}$ \\
    \textbf{IX}. SimNet & $71.05$ & $69.26$ & $70.78$ & $74.50$  & \textbf{XIX}. CIG-Sim\&Siam-GCN-BERT$^{g}$ & $\mathbf{84.68}$ & $82.60$ & $89.56$ & $89.97$ \\ \cline{1-5}
    \textbf{X}. BERT fine-tuning & $81.30$ & $79.20$ & $86.64$ & $87.08$  & \textbf{XX}. CIG-Sim\&Siam-GCN-Sim$^{g}$\&BERT$^{g}$ & $84.61$ & $82.59$ & $89.47$ & $89.71$ \\
    \bottomrule
  \end{tabular}}
  \vspace{-5mm}
\end{table*}

\textbf{Baselines}.
We test the following baselines:
\begin{itemize}
\item \textit{Matching by representation-focused or interaction-focused deep neural network models}: DSSM \cite{huang2013learning}, C-DSSM \cite{shen2014learning}, DUET \cite{mitra2017learning}, MatchPyramid \cite{pang2016text}, ARC-I \cite{hu2014convolutional}, ARC-II \cite{hu2014convolutional}.
We use the implementations from MatchZoo~\cite{fan2017matchzoo} for the evaluation of these models.
\item \textit{Matching by term-based similarities}: BM25 \cite{robertson2009probabilistic}, LDA \cite{blei2003latent} and SimNet (which is extracting the five text-pair similarities mentioned in Sec.~\ref{sec:textmatch} and classifying by a multi-layer feedforward neural network).
\item \textit{Matching by a large-scale pre-training language model}: BERT \cite{devlin2018bert}.
\end{itemize}

Note that we focus on the capability of long text matching. Therefore, we do not use any short text information, such as titles, in our approach or in any baselines. 
In fact, the ``relationship" between two documents is not limited to "whether the same event or not". 
Our algorithm is able to identify a general relationship between documents, e.g., whether two episodes are from the same season of a TV series. The definition of the relationship (e.g., same event/story, same chapter of a book) is solely defined and supervised by the labeled training data. For these tasks, the availability of other information such as titles can not be assumed.

As shown in Table~\ref{tab:experiments}, we evaluate different variants of our own model to show the effect of different sub-modules.
In model names, ``CIG'' means that in CIG, we directly use keywords as concepts without community detection, whereas
``CIG$_{cd}$'' means that each concept vertex in the CIG contains a set of keywords grouped via community detection.
To generate the matching vector on each vertex,
``Siam'' indicates the use of Siamese encoder, while ``Sim'' indicates the use of term-based similarity encoder, as shown in Fig.~\ref{fig:System}.
``GCN'' means that we convolve the local matching vectors on vertices through GCN layers.
Finally, ``BERT$^g$'' or ``Sim$^g$'' indicates the use of additional global features given by BERT or the five term-based similarity metrics mentioned in Sec.~\ref{sec:textmatch}, appended to the graphically merged matching vector $\mathbf m_{AB}$, for final classification.

\textbf{Implementation Details}.
We use Stanford CoreNLP~\cite{manning2014stanford} for word segmentation (on Chinese text) and named entity recognition.
For Concept Interaction Graph construction with community detection, we set the minimum community size (number of keywords contained in a concept vertex) to be 2, and the maximum size to be 6.

Our neural network model consists of word embedding layer, Siamese encoding layer, Graph transformation layers, and classification layer.
For embedding, we load the pre-trained word vectors and fix it during training. The embeddings of out of vocabulary words are set to be zero vectors.
For the Siamese encoding network, we use 1-D convolution with number of filters $32$, followed by an ReLU layer and Max Pooling layer.
For graph transformation, we utilize 2 layers of GCN \cite{kipf2016semi} for experiments on the CNSS dataset, and 3 layers of GCN for experiments on the CNSE dataset.
When the vertex encoder is the five-dimensional features, we set the output size of GCN layers to be 16. When the vertex encoder is the Siamese network encoder, we set the output size of GCN layers to be 128 except the last layer. For the last GCN layer, the output size is always set to be $16$.
For the classification module, it consists of a linear layer with output size $16$, an ReLU layer, a second linear layer, and finally a Sigmoid layer.
Note that this classification module is also used for the baseline method SimNet.

As we mentioned in Sec.~\ref{sec:intro}, our code and datasets have been open sourced.
We implement our model using PyTorch  1.0 \cite{paszke2017pytorch}. The experiments without BERT are carried out on an MacBook Pro with a 2 GHz Intel Core i7 processor and 8 GB memory.
We use L2 weight decay on all the trainable variables, with parameter $\lambda = 3 \times 10^{-7}$. The dropout rate between every two layers is $0.1$. We apply gradient clipping with maximum gradient norm $5.0$.
We use the ADAM optimizer \cite{kingma2014adam} with $\beta_1 = 0.8,\ \beta_2 = 0.999,\ \epsilon = 10^{−8}$. We use a learning rate warm-up scheme with an inverse exponential increase from 0.0 to 0.001 in the first 1000 steps, and then maintain a constant learning rate for the remainder of training. For all the experiments, we set the maximum number of training epochs to be 10.

\subsection{Results and Analysis}
\label{subsec:performance}

Table~\ref{tab:experiments} summarizes the performance of all the compared methods on both datasets.
Our model achieves the best performance on both two datasets and significantly outperforms all other methods.
This can be attributed to two reasons. First, as the input of article pairs are re-organized into Concept Interaction Graphs, the two documents are aligned along the corresponding semantic units for easier concept-wise comparison. Second, our model encodes local comparisons around different semantic units into local matching vectors, and aggregate them via graph convolutions, taking semantic topologies into consideration. Therefore, it solves the problem of matching documents via divide-and-conquer, which is suitable for handling long text. 

\textbf{Impact of Graphical Decomposition.}
Comparing method XI with methods I-VI in Table~\ref{tab:experiments}, they all use the same word vectors and use neural networks for text encoding. The key difference is that our method XI compares a pair of articles over a CIG in per-vertex decomposed fashion. We can see that the performance of method XI is significantly better than methods I-VI. Similarly, comparing our method XIV with methods VII-IX, they all use the same term-based similarities. However, our method achieves significantly better performance by using graphical decomposition. Therefore, we conclude that graphical decomposition can greatly improve long text matching performance.

Note that the deep text matching models I-VI lead to bad performance, because they were invented mainly for sequence matching and can hardly capture meaningful semantic interactions in article pairs. When the text is long, it is hard to get an appropriate context vector representation for matching. For interaction-focused neural network models, most of the interactions between words in two long articles will be meaningless. 

\textbf{Impact of Graph Convolutions.} Compare methods XII and XI, and compare methods XV and XIV. We can see that incorporating GCN layers has significantly improved the performance on both datasets. 
Each GCN layer updates the hidden vector of each vertex by integrating the vectors from its neighboring vertices. Thus, the GCN layers learn to graphically aggregate local matching features into a final result.

\textbf{Impact of Community Detection.}
By comparing methods XIII and XII, and comparing methods XVI and XV, we observe that using community detection, such that each concept is a set of correlated keywords instead of a single keyword, leads to slightly worse performance. This is reasonable, as using each keyword directly as a concept vertex provides more anchor points for article comparison . However, community detection can group highly coherent keywords together and reduces the average size of CIGs from $30$ to $13$ vertices. This helps to reduce the total training and testing time of our models by as much as $55$\%. Therefore, one may choose whether to apply community detection to trade accuracy off for speedups.

\textbf{Impact of Multi-viewed Matching.}
Comparing methods XVII and XV, we can see that the concatenation of different graphical matching vectors (both term-based and Siamese encoded features) can further improve performance. This demonstrates the advantage of combining multi-viewed matching vectors.

\textbf{Impact of Added Global Features.} Comparing methods XVIII, XIX, XX with method XVII, we can see that adding more global features, such as global similarities (Sim$^g$) and/or global BERT encodings (BERT$^g$) of the article pair, can hardly improve performance any further. This shows that graphical decomposition and convolutions are the main factors that contribute to the performance improvement. 
Since they already learn to aggregate local comparisons into a global semantic relationship, additionally engineered global features cannot help. 

\textbf{Model Size and Parameter Sensitivity}:
Our biggest model without BERT is XVIII, which contains only $\sim$34K parameters. In comparison, BERT contains 110M-340M parameters. However, our model significantly outperforms BERT. 

We tested the sensitivity of different parameters in our model. We found that 2 to 3 layers of GCN layers gives the best performance. Further introducing more GCN layers does not improve the performance, while the performance is much worse with zero or only one GCN layer. Furthermore, in GCN hidden representations of a size between 16 and 128 yield good performance. Further increasing this size does not show obvious improvement.

For the optional community detection step in CIG construction, we need to choose the minimum size and the maximum size of communities. We found that the final performance remains similar if we vary the minimum size from 2$\sim$3 and the maximum size from 6$\sim$10. This indicates that our model is robust and insensitive to these parameters.

\textbf{Time complexity.}
For keywords of news articles, in real-world industry applications, they are usually extracted in advance by highly efficient off-the-shelf tools and pre-defined vocabulary.
For CIG construction, let $N_s$ be the number of sentences in two documents, $N_w$ be the number of unique words in documents, and $N_k$ represents the number of unique keywords in a document. Building keyword graph requires $\mathcal{O}(N_s N_k + N_w^2)$ complexity \cite{sayyadi2013graph}, and betweenness-based community detection requires $\mathcal{O}(N_k^3)$. The complexity of sentence assignment and weight calculation is $\mathcal{O}(N_s N_k + N_k^2)$.
For graph classification, our model size is not big and can process document pairs efficiently.

\section{Related Work}
\label{sec:related}

\textbf{Graphical Document Representation.}
A majority of existing works can be generalized into four categories: word graph, text graph, concept graph, and hybrid graph.
Word graphs use words in a document as vertices, and construct edges based on syntactic analysis~\cite{leskovec2004learning}, co-occurrences~\cite{zhang2018multiresolution,rousseau2013graph,nikolentzos2017shortest} or preceding relation~\cite{schenker2003clustering}.
Text graphs use sentences, paragraphs or documents as vertices, and establish edges by word co-occurrence, location~\cite{mihalcea2004textrank}, text similarities~\cite{putra2017evaluating}, or hyperlinks between documents~\cite{page1999pagerank}.
Concept graphs link terms in a document to real world concepts based on knowledge bases such as DBpedia \cite{auer2007dbpedia}, and construct edges based on syntactic/semantic rules.
Hybrid graphs \cite{rink2010learning,baker2017graph} consist of different types of vertices and edges.

\textbf{Text Matching.}
Traditional methods represent a text document as vectors of bag of words (BOW), term frequency inverse document frequency (TF-IDF), LDA \cite{blei2003latent} and so forth, and calculate the distance between vectors. However, they cannot capture the semantic distance and usually cannot achieve good performance.

In recent years, different neural network architectures have been proposed for text pair matching tasks.
For representation-focused models, they usually transform text pairs into context representation vectors through a Siamese neural network, followed by a fully connected network or score function which gives the matching result based on the context vectors \cite{qiu2015convolutional,wan2016deep,liu2018matching,mueller2016siamese,severyn2015learning}.
For interaction-focused models, they extract the features of all pair-wise interactions between words in text pairs, and aggregate the interaction features by deep networks to give a matching result \cite{hu2014convolutional,pang2016text}.
However, the intrinsic structural properties of long text documents are not fully utilized by these neural models. Therefore, they cannot achieve good performance for long text pair matching.

There are also research works which utilize knowledge \cite{wu2018knowledge}, hierarchical property \cite{jiang2019semantic} or graph structure \cite{nikolentzos2017shortest,paul2016efficient} for long text matching. In contrast, our method represents documents by a novel graph representation and combines the representation with GCN.

Finally, pre-training models such as BERT \cite{devlin2018bert} can also be utilized for text matching. However, the model is of high complexity and is hard to satisfy the speed requirement in real-world applications.

\textbf{Graph Convolutional Networks.} We also contributed to the use of GCNs to identify the relationship between \emph{a pair of} graphs, whereas previously, different GCN architectures have mainly been used for completing missing attributes/links \cite{kipf2016semi, defferrard2016convolutional} or for node clustering or classification \cite{hamilton2017representation}, but all within the context of a \emph{single} graph, e.g., a knowledge graph, citation network or social network.
In this work, the proposed Concept Interaction Graph takes a simple approach to represent a document by a weighted undirected graph, which essentially helps to decompose a document into subsets of sentences, each subset focusing on a different sub-topic or concept.

\section{Conclusion}
\label{sec:conclude}
We propose the \textit{Concept Interaction Graph} 
to organize documents into a graph of concepts, and introduce a divide-and-conquer approach to matching a pair of articles based on graphical decomposition and convolutional aggregation.  
We created two new datasets for long document matching with the help of professional editors, consisting of about 60K pairs of news articles, on which we have performed extensive evaluations. In the experiments, our proposed approaches significantly outperformed an extensive range of state-of-the-art schemes, including both term-based and deep-model-based text matching algorithms.
Results suggest that the proposed graphical decomposition and the structural transformation by GCN layers are critical to the performance improvement in matching article pairs.

\clearpage
\bibliography{main}
\bibliographystyle{acl_natbib}

\end{document}